\documentclass{article}

\usepackage{arxiv}

\usepackage[utf8]{inputenc} 
\usepackage[T1]{fontenc}    
\usepackage{hyperref}       
\usepackage{url}            
\usepackage{booktabs}       
\usepackage{amsfonts}       
\usepackage{nicefrac}       
\usepackage{microtype}      

\usepackage[obeyDraft]{todonotes}\presetkeys{todonotes}{inline}{}
\usepackage{pgfplots}\usetikzlibrary{plotmarks}\pgfplotsset{compat=1.14}

\usepackage[ruled,linesnumbered,vlined]{algorithm2e}

\usepackage{amsmath}
\usepackage{amssymb}
\usepackage{booktabs}
\usepackage{comment}
\usepackage{enumitem}
\usepackage{multirow}
\usepackage{url}
\usepackage{thm-restate}
\usepackage{oz}
\usepackage{listings}

\title{The pyglaf argumentation reasoner (ICCMA2021)}

\author{
  Mario Alviano \\
  Department of Mathematics and Computer Science\\
  University of Calabria\\
  Rende (CS), IT 87036 \\
  \texttt{mario.alviano@unical.it} \\
}

\begin{document}
\maketitle

\begin{abstract}
The \textsc{pyglaf} reasoner takes advantage of circumscription to solve computational problems of abstract argumentation frameworks.
In fact, many of these problems are reduced to circumscription by means of linear encodings, and a few others are solved by means of a sequence of calls to an oracle for circumscription.
Within \textsc{pyglaf}, Python is used to build the encodings and to control the execution of the external circumscription solver, which extends the SAT solver \textsc{glucose} and implements algorithms taking advantage of unsatisfiable core analysis and incremental computation.
\end{abstract}

\section{Introduction}
Circumscription \cite{DBLP:journals/ai/McCarthy80} is a nonmonotonic logic formalizing common sense reasoning by means of a second order semantics, which essentially enforces to minimize the extension of some predicates.
With a little abuse on the definition of circumscription, the minimization can be imposed on a set of literals, so that a set of negative literals can be used to encode a maximization objective function.
Since many semantics of abstract argumentation frameworks are based on a preference relation that essentially amount to inclusion relationships, \textsc{pyglaf} (\url{http://alviano.com/software/pyglaf/}) uses circumscription as a target language to solve computational problems of abstract argumentation frameworks.

\textsc{pyglaf} \cite{DBLP:conf/iclp/Alviano17} is implemented in Python and uses \textsc{circumscriptino} (\url{http://alviano.com/software/circumscriptino/}), a circumscription solver extending the SAT solver \textsc{glucose} \cite{DBLP:conf/ijcai/AudemardS09}.
Linear reductions are used for all semantics \cite{DBLP:conf/aiia/Alviano17}.
For the ideal extension, the reduction requires the union of all admissible extensions of the input graph;
such a set is computed by means of iterative calls to \textsc{circumscriptino}.
The communication between \textsc{pyglaf} and \textsc{circumscriptino} is handled in the simplest possible way, that is, via stream processing.
This design choice is principally motivated by the fact that the communication is often minimal, limited to a single invocation of the circumscription solver.

The reasoner supports the following ICCMA 2021 problems:
CE-CO, CE-PR, CE-SST, CE-ST, DC-CO, DC-PR, DC-SST, DC-ST, DC-STG, DS-CO, DS-ID, DS-PR, DS-SST, DS-ST, DS-STG, SE-CO, SE-ID, SE-PR, SE-SST, SE-ST, and SE-STG.
Support for CE-STG is a work in progress.

\section{Circumscription}

Let $\mathcal{A}$ be a fixed, countable set of \emph{atoms} including $\bot$.
A \emph{literal} is an atom possibly preceded by the connective $\neg$.
For a literal $\ell$, let $\overline{\ell}$ denote its \emph{complementary literal}, that is, $\overline{p} = \neg p$ and $\overline{\neg p} = p$ for all $p \in \mathcal{A}$;
for a set $L$ of literals, let $\overline{L}$ be $\{\overline{\ell} \mid \ell \in L\}$.
\emph{Formulas} are defined as usual by combining atoms and the connectives $\neg$, $\wedge$, $\vee$, $\rightarrow$, $\leftrightarrow$.
A \emph{theory} is a set $T$ of formulas including $\neg \bot$;
the set of atoms occurring in $T$ is denoted by $\mathit{atoms}(T)$.
An \emph{assignment} is a set $A$ of literals such that $A \cap \overline{A} = \emptyset$.
An \emph{interpretation} for a theory $T$ is an assignment $I$ such that $(I \cup \overline{I}) \cap \mathcal{A} = \mathit{atoms}(T)$.
Relation $\models$ is defined as usual.
$I$ is a \emph{model} of a theory $T$ if $I \models T$.
Let $\mathit{models}(T)$ denote the set of models of $T$.

\emph{Circumscription} applies to a theory $T$ and a set $P$ of literals subject to minimization.
Formally, relation $\leq^{P}$ is defined as follows:
for $I,J$ interpretations of $T$, $I \leq^{P} J$ if $I \cap P \subseteq J \cap P$.
$I \in \mathit{models}(T)$ is a \emph{preferred model} of $T$ with respect to $\leq^{P}$ if there is no $J \in \mathit{models}(T)$ such that $I \not\leq^{P} J$ and $J \leq^{P} I$.
Let $\mathit{CIRC}(T,P)$ denote the set of preferred models of $T$ with respect to $\leq^{P}$.

\section{From Argumentation Frameworks to Circumscription}\label{sec:red}

An \emph{abstract argumentation framework} (AF) is a directed graph $G$ whose nodes $\mathit{arg}(G)$ are arguments, and whose arcs $\mathit{att}(G)$ represent an attack relation.
An \emph{extension} $E$ is a set of arguments.
The \emph{range} of $E$ in $G$ is $E^+_G := E \cup \{x \mid \exists yx \in \mathit{att}(G)$ with $y \in E\}$.
In the following, the semantics of ICCMA'17 are characterized by means of circumscription.

For each argument $x$, an atom $a_x$ is possibly introduced to represent that $x$ is attacked by some argument that belongs to the computed extension $E$, and an atom $r_x$ is possibly introduced to enforce that $x$ belongs to the range $E^+_G$:
\begin{align}
    \label{eq:attacked}
    \mathit{attacked(G)} :={}& \left\{a_x \leftrightarrow \bigvee_{yx \in \mathit{att}(G)}{y} \  \middle|\ x \in \mathit{arg}(G)\right\}\\
    \label{eq:range}
    \mathit{range(G)} :={}& \left\{r_x \rightarrow x \vee \bigvee_{yx \in \mathit{att}(G)}{y} \ \middle|\ x \in \mathit{arg}(G)\right\}
\end{align}
The following set of formulas characterize semantics not based on preferences:
\begin{align}
    \label{eq:conflict-free}
    \mathit{conflict\textit{-}free}(G) :={} & \{\neg \bot\} \cup \{\neg x \vee \neg y \mid xy \in \mathit{att}(G)\} \\
    \label{eq:admissible}
    \mathit{admissible}(G) :={} & \mathit{conflict\textit{-}free}(G)\!\cup\!\mathit{attacked}(G) \cup \{x \rightarrow a_y \mid yx \in \mathit{att}(G)\} \\
    \label{eq:complete}
    \mathit{complete}(G) :={} & \mathit{admissible}(G) \cup \left\{\left(\bigwedge_{yx \in \mathit{att}(G)}{a_y}\right) \rightarrow x \ \middle|\ x \in \mathit{arg}(G)\right\}\\
    \label{eq:stable}
    \mathit{stable}(G) :={} & \mathit{complete}(G) \cup \mathit{range}(G) \cup \{r_x \mid x \in \mathit{arg}(G)\}
\end{align}
Note that in (\ref{eq:admissible}) truth of an argument $x$ implies that all arguments attacking $x$ are actually attacked by some true argument.
In (\ref{eq:complete}), instead, whenever all attackers of an argument $x$ are attacked by some true argument, argument $x$ is forced to be true.
Finally, in (\ref{eq:stable}) all atoms of the form $r_x$ are forced to be true, so that the range of the computed extension has to cover all arguments.

The ideal semantic is defined as follows (Proposition~3.6 by \cite{DBLP:journals/argcom/Caminada11}):
Let $X$ be the set of admissible extensions of $G$ that are not attacked by any admissible extensions, that is, $X := \{E \in \mathit{models}(\mathit{admissible}(G)) \mid \nexists E' \in \mathit{models}(\mathit{admissible}(G))$ such that $yx \in \mathit{att}(G), x \in E, y \in E'\}$.
$E$ is the ideal extension of $G$ if $E \in X$, and there is no $E' \in X$ such that $E' \supseteq E$.

All semantics of ICCMA'19 are characterized in circumscription as follows:
\begin{align}
    \label{eq:co}
    \mathit{co}(G) :={} & \mathit{CIRC}(\mathit{complete}(G), \emptyset) \\
    \label{eq:st}
    \mathit{st}(G) :={} & \mathit{CIRC}(\mathit{stable}(G), \emptyset)\\ 
    \label{eq:gr}
    \mathit{gr}(G) :={} & \mathit{CIRC}(\mathit{complete}(G), \mathit{arg}(G))\\
    \label{eq:pr}
    \mathit{pr}(G) :={} & \mathit{CIRC}(\mathit{complete}(G), \overline{\mathit{arg}(G)})\\
    \label{eq:sst}
    \mathit{sst}(G) :={} & \mathit{CIRC}(\mathit{complete}(G) \cup \mathit{range}(G), \{\neg r_x \mid x \in \mathit{arg}(G)\})\\
    \label{eq:stg}
    \mathit{stg}(G) :={} & \mathit{CIRC}(\mathit{conflict\textit{-}free}(G) \cup \mathit{range}(G), \{\neg r_x \mid x \in \mathit{arg}(G)\})\\
    \label{eq:id}
    \mathit{id}(G,U) :={} & \mathit{CIRC}(\mathit{admissible}(G) \cup \overline{\mathit{arg}(G) \setminus Y},\ \overline{Y})
\end{align}
where in (\ref{eq:id}) $U$ is the union of all admissible extensions of $G$, and $Y$ is $U \setminus \{x \mid \exists yx \in \mathit{att}(G),\ y \in U\}$.

\section{Implementation}

Abstract argumentation frameworks can be encoded in trivial graph format (\textsc{tgf}) as well as in aspartix format (\textsc{apx}).
The following data structures are populated during the parsing of the input graph $G$:
a list \texttt{arg} of the arguments in $\mathit{arg}(G)$;
a dictionary \texttt{argToIdx}, mapping each argument $x$ to its position in \texttt{arg};
a dictionary \texttt{att}, mapping each argument $x$ to the set $\{y \mid xy \in \mathit{att}(G)\}$;
a dictionary \texttt{attR}, mapping each argument $x$ to the set $\{y \mid yx \in \mathit{att}(G)\}$.
Within these data structures, theories (\ref{eq:co})--(\ref{eq:id}) are constructed in amortized linear time.
Single extension computation and extension enumeration is then demanded to the underlying circumscription solver \cite{DBLP:journals/tplp/Alviano17}.



The union $U$ of all admissible extensions is computed by iteratively asking to \textsc{circumscriptino} to compute an admissible extension that maximize the accepted arguments not already in $U$, so to expand $U$ as much as possible at each iteration.


For complete, stable, and preferred extensions, credulous acceptance is addressed by checking consistency of the theory extended with the query argument.
Similarly, skeptical acceptance is addressed by adding the complement of the query argument for complete, and stable extensions.
Grounded and ideal extensions are unique, and therefore credulous acceptance is addressed by checking the presence of the query argument in the computed extension.
Actually, for the ideal extension, a negative answer is possibly produced already if the query argument is not part of the union of all admissible extensions.
The remaining acceptance problems are addressed by a recent algorithm for query answering in circumscription \cite{DBLP:conf/ijcai/Alviano18}.
In a nutshell, given a query atom $q$ and a circuscribed theory, the computational problem amounts to search for a model of the circumscribed theory that contains the query atom.
The algorithm implemented in \textsc{circumscriptino} searches for a classical model of the theory that contains the query atom, and checks that no more preferred model not containing the query atom exists.
In this way, queries are possibly answered without computing any optimal model.

Counting of extensions is implemented by enumeration, with no particular optimization.

\section*{Acknowledgments}

This work was partially supported by the projects PON-MISE MAP4ID (CUP B21B19000650008) and PON-MISE S2BDW (CUP B28I17000250008), by the LAIA lab (part of the SILA labs) and by GNCS-INdAM.

\bibliographystyle{unsrt}  
\bibliography{bibtex} 
\end{document}